\documentclass[11pt,a4paper]{article}
\usepackage[hyperref]{acl2019}
\usepackage{times}
\usepackage{latexsym}
\usepackage{multirow}
\usepackage{amssymb,amsmath,amsthm,amsfonts}
\usepackage{times}
\usepackage{latexsym}
\usepackage{multicol}
\usepackage{multirow}
\usepackage{algorithm}
\usepackage[noend]{algpseudocode}
\usepackage[colorinlistoftodos]{todonotes}
\usepackage{comment}
\usepackage{enumitem}
\usepackage{breqn}
\usepackage{tabularx}
\usepackage{arydshln}
\usepackage{graphicx}
\usepackage{makecell}
\usepackage{caption}
\usepackage{subcaption}
\usepackage{url}
\usepackage{color,soul}
\usepackage{xcolor}
\usepackage{bbding}
\usepackage{appendix}
\usepackage{multirow}
\usepackage{float}

\DeclareMathOperator*{\id}{\mathbf{I}}

\aclfinalcopy 
\def\aclpaperid{9} 

\title{Conceptor Debiasing of Word Representations Evaluated on WEAT}

\author{
  Saket Karve \\
  \And
  Lyle Ungar \\
  Department of Computer \& Information Science\\
  University of Pennsylvania\\
  Philadelphia, PA 19104 \\
  \texttt{\{saketk,ungar,joao\} @cis.upenn.edu}
  \And
  Jo\~{a}o Sedoc \\
}

\begin{document}

\maketitle

\begin{abstract}
Bias in word embeddings such as Word2Vec has been widely investigated, and many efforts made to remove such bias.  We show how to use {\em conceptors debiasing} to post-process both traditional and contextualized word embeddings. Our conceptor debiasing can simultaneously remove racial and gender biases and,  unlike standard debiasing methods, can make effective use of heterogeneous lists of biased words. We show that conceptor debiasing diminishes racial and gender bias of word representations as measured using the Word Embedding Association Test (WEAT) of \citet{caliskan2017semantics}.
\end{abstract}

\maketitle

\section{Introduction}

Word embeddings capture distributional similarities and thus inherit demographic stereotypes~\citep{bolukbasi2016man}. Such embedding biases tend to track statistical regularities such as the percentage of people with a given occupation~\citep{garg2018} but sometimes deviate from them~\citep{bhatia2017semantic}. Recent work has shown that gender bias exists in contextualized embeddings~\citep{wang2019genderbiaselmo,may2019measuring}.

\begin{figure}[ht!]
\centering
\begin{subfigure}{0.47\textwidth}
  \centering
  \includegraphics[width=1\linewidth]{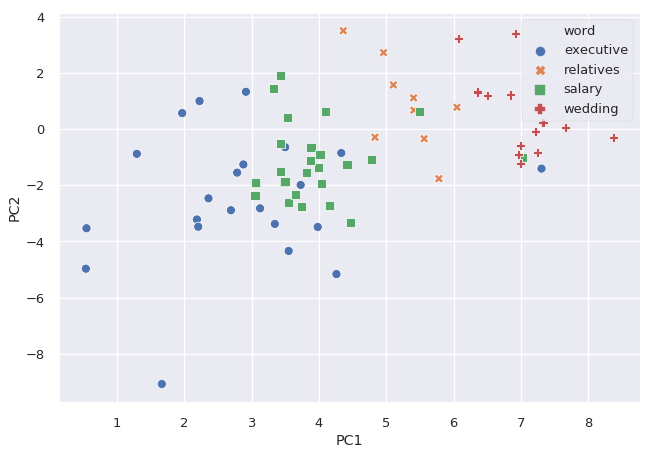}
  \caption{The original space}
  \label{fig:pcbertcareerhomeontogendernames}
\end{subfigure} 
\\
\begin{subfigure}{.47\textwidth}
  \centering
  \includegraphics[width=1\linewidth]{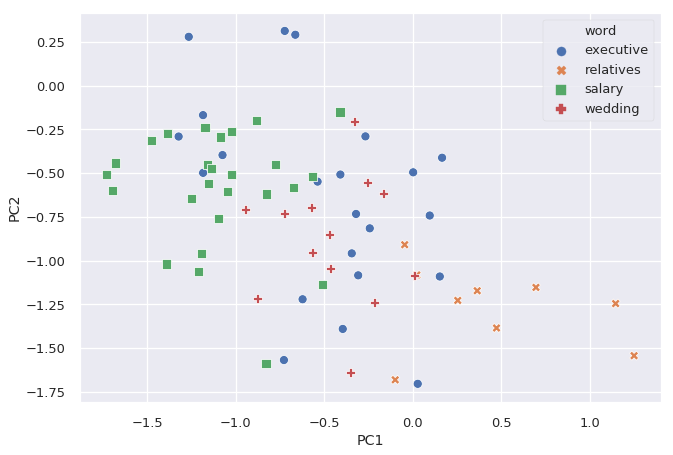}
  \caption{After applying the debiasing conceptor}
  \label{ffig:debiaspcbertcareerhomeontogendernames}
\end{subfigure}
\caption{BERT word representations of the union of the  set  of  contextualized  word  representations  of {\it \textcolor[rgb]{.84,.48,.05}{relatives}, \textcolor[rgb]{0,0,1}{executive}, \textcolor[rgb]{0.7,0,.0}{wedding}, \textcolor[rgb]{0,0.4,0}{salary}} projected on to the first two principal components of the WEAT gender first names, which capture the primary component of gender. Note how the debiasing conceptor collapses {\it relatives} and {\it wedding}, and {\it executive} and {\it salary} once the bias is removed. }
\label{fig:debiasingconceptor}
\end{figure}

Here, we provide a quantitative analysis of bias in traditional and contextual word embeddings and introduce a method of mitigating bias (i.e., debiasing) using {\em the debiasing conceptor}, a clean mathematical representation of subspaces that can be operated on and composed by logic-based  manipulations~\citep{Jaeger2014}.  
Specifically, conceptor negation is a soft damping of the principal components of the target subspace (e.g., the subset of words being debiased)~\citep{Liu2019} (See Figure \ref{fig:debiasingconceptor}.)

Key to our method is how it treats word-association lists (sometimes called target lists), which define the bias subspace.  These lists include pre-chosen words associated with a target demographic group (often referred to as a ``protected class''). For example, {\it he / she} or {\it Mary / John} have been used for gender~\citep{bolukbasi2016man}. More generally, conceptors can combine multiple subspaces defined by word lists. Unlike most current methods, conceptor debiasing uses a soft, rather than a hard projection.

We test the debiasing conceptor on a range of traditional and contextualized word embeddings\footnote{Previous work has shown that debiasing methods can have different effects on different word embeddings~\citep{kiritchenko2018examining}.
} and examine whether they remove stereotypical demographic biases. All tests have been performed on English word embeddings.

This paper contributes the following:
\begin{itemize}[noitemsep,nolistsep]
    \item Introduces {\em debiasing conceptors} along with a formal definition and mathematical relation to the Word Embedding Association Test.
    \item Demonstrates the effectiveness of the debiasing conceptor on both traditional and contextualized word embeddings.
\end{itemize}

\section{Related Work}
NLP has begun tackling the problems that inhibit the achievement of fair and ethical AI~\citep{hovy2016social,friedler2016possibility}, in part by developing techniques for mitigating demographic biases in models.
In brief, a {\it demographic bias} is a difference in model output based on gender (either of the data author or of the content itself) or selected demographic dimension (``protected class") such as race. 
Demographic biases manifest in many ways, ranging from disparities in tagging and classification accuracy depending on author age and gender~\cite{hovy2015demographic,dixon2018measuring}, to over-amplification of demographic differences in language generation~\citep{yatskar2016situation,zhao2017men}, to diverging implicit associations between words or concepts within embeddings or language models~\citep{bolukbasi2016man,rudinger2018gender}. 

Here, we are concerned with the societal bias towards protected classes that manifests in prejudice and stereotypes~\citep{bhatia2017semantic}. \citet{greenwald1995implicit};  implicit attitudes such that ``introspectively unidentified (or inaccurately identified) traces of past experience that mediate favorable or unfavorable feeling, thought, or action toward social objects.''  Bias is often quantified in people using the Implicit Association Test (IAT)~\citep{greenwald1998measuring}. The IAT records subjects response times when asked to pair
two concepts. Smaller response times occur in concepts subjects perceive to be similar versus pairs of
concepts they perceive to be different. A well known example is where subjects were asked to associate black and white names with ``pleasant'' and ``unpleasant'' words. A significant racial bias has been found in many populations.
Later, \citet{caliskan2017semantics} formalized the Word Embedding Association Test (WEAT), which replaces reaction time with word similarity to give a bias measure that does not require use of human subjects. \citet{may2019measuring} extended WEAT to the Sentence Embedding Association Test (SEAT); however, in this paper we instead use token-averaged representations over a corpus.

\paragraph*{Debiasing Embeddings. }
The simplest way to remove bias is to project out a bias direction. For example, \citet{bolukbasi2016man} identify a ``gender subspace'' using lists of gendered words and then remove the first principal component of this subspace. \citet{wang2019genderbiaselmo} used both data augmentation and debiasing of \citet{bolukbasi2016man} to mitigate bias found in ELMo and showed improved performance on coreference resolution. Our work is complementary, as debiasing conceptors can be used in place of hard-debiasing. 

\citet{bolukbasi2016man} also examine a soft debiasing method, but find that it does not perform well. In contrast, our debiasing conceptor does a successful soft damping of the relevant principal components. To understand why, we first introduce the conceptor method for capturing the ``bias subspaces'', next formalize bias, and then show WEAT in matrix notation.

\subsection{Conceptors}

As in \citet{bolukbasi2016man}, our aim is to identify the ``bias subspace'' using a set of target words, $\mathcal{Z}$ and $Z$ is their corresponding word embeddings. A conceptor matrix, $C$, is a regularized identity map (in our case, from the original word embeddings to their biased versions) that minimizes
\begin{equation}
\label{matconceptor}
\|Z - CZ\|_F^2 + \alpha^{-2}\|C\|_{F}^2.
\end{equation}
where $\alpha^{-2}$ is a scalar parameter.\footnote{Note that the conceptor and WEAT literature disagree on notation and we follow WEAT. In conceptor notation, the matrix $Z$ would be denoted as $X$.} 

To describe matrix conceptors, we draw heavily on  \citep{Jaeger2014,He2018,Liu2019,liu2019continual}. 
$C$ has a closed form solution: 

\begin{equation}
\label{conceptorsolution}
C = \frac{1}{k} Z Z^{\top} (\frac{1}{k} Z Z^{\top}+\alpha^{-2} I)^{-1}.
\end{equation}
Intuitively, $C$ is a soft projection matrix on the linear subspace where the word embeddings $Z$ have the highest variance. Once $C$ has been learned, it can be `negated' by subtracting it from the identity matrix and then applied to any word embeddings to shrink the bias directions.

Conceptors can represent laws of Boolean logic, such as NOT $\neg$, AND $\wedge$ and OR $\vee$. For two conceptors $C$ and $B$, we define the following operations:
\begin{align}
\neg C:=& \id-C, \label{eqn:negconcept} \\
C\wedge B:=&  (C^{-1} + B^{-1} - \id)^{-1} \\
C \vee B:=&\neg(\neg C \wedge \neg B) 
\end{align}
Among these Boolean operations, two are critical for this paper: the NOT operator for debiasing, and the OR operation $\vee$ for multi-list (or multi-category) debiasing. It can be shown that if $C$ and $B$ are of equal sizes, then $C \vee B$ is the conceptor computed from the union of the two sets of sample points from which $C$ and $B$ are computed
\cite{Jaeger2014}; this is not true if they are of different sizes.

\paragraph{Negated Conceptor.}
Given that the conceptor, $C$, represents the subspace of maximum bias, we want to apply the negated conceptor, NOT $C$ (see Equation \ref{eqn:negconcept}) to an embedding space and remove its bias. We call NOT $C$ the {\it debiasing conceptor}. More generally, if we have $K$ conceptors, $C_i$ derived from $K$ different word lists, we call NOT $(C_1 \vee ... \vee C_K)$ a debiasing conceptor. 
The negated conceptor matrix has been used in the past on a complete vocabulary 
to increase the semantic richness of its word embeddings; \citet{Liu2018correcting} showed that the negated conceptor gave better performance on semantic similarity and downstream tasks  than the hard debiasing method of \citet{Mu2018}. 

As shown in \citet{Liu2018correcting}, the negated conceptor approach does a soft debiasing by shrinking each principal component of the covariance matrix of the target word embeddings $Z Z^{\top}$. The shrinkage is a function of the conceptor hyper-parameter $\alpha$ and the singular values $\sigma_i$ of $Z Z^{\top}$:
$\frac{\alpha^{-2}}{\sigma_{i}+\alpha^{-2}} $ .

\section{Formalizing Bias}
We follow the formal definition of \citet{lu2018gender}, where 
given a class of word sets $\mathcal{D}$ and a scoring function
$s$, the bias of $s$ under the concept(s) tested by $\mathcal{D}$, written $\mathcal{B}_s(\mathcal{D})$, is the expected difference in scores assigned to expected absolute bias across class members,
\[
\mathcal{B}_s(\mathcal{D})  \triangleq \mathbb{E}_{D \in \mathcal{D}}|\mathcal{B}_s(D)|.
\]
\noindent
This naturally gives rise to a large set of concepts and scoring functions. 

\subsection{Word Embedding Association Test}
\label{sec:motivationtheory}

The Word Embeddings Association Test (WEAT), as proposed by \citet{caliskan2017semantics}, is a statistical test analogous to the Implicit Association Test (IAT)~\citep{greenwald1998measuring} which helps quantify human biases in textual data. WEAT uses the cosine similarity between word embeddings, which is analogous to the reaction time when subjects are asked to pair two concepts they find similar in the IAT.  WEAT considers two sets of target words and two sets of attribute words of equal size. The null hypothesis is that there is no difference between the two sets of target words and the sets of attribute words in terms of their relative similarities measured as the cosine similarity between the embeddings. For example, consider the target sets as words representing \textit{Career} and \textit{Family} and let the two sets of attribute words be \textit{Male} and \textit{Female}, in that order. The null hypothesis states that \textit{Career} and \textit{Family} are equally similar (mathematically, in terms of the mean cosine similarity between the word representations) to each of the words in the \textit{Male} and \textit{Female} word lists. 

The WEAT test statistic measures the differential association of the two sets of target words with the attribute. The ``effect size'' is a normalized measure of how separated the two distributions are.

To ground this, we cast WEAT in our formulation where $\mathcal{X}$ and $\mathcal{Y}$ are two sets of target
words, (concretely, $\mathcal{X}$ might be \textit{Career} words and $\mathcal{Y}$ \textit{Family} words) and $\mathcal{A}$, $\mathcal{B}$ are two sets of attribute words ($\mathcal{A}$ might be \textit{female} names and $\mathcal{B}$ \textit{male} names) assumed to associate with the bias concept(s). WEAT is then~\footnote{We assume that there is no overlap between any of the sets $\mathcal{X}$, $\mathcal{Y}$, $\mathcal{A}$, and $\mathcal{B}$.} 
\begin{align*}
s(\mathcal{X}, &\mathcal{Y}, \mathcal{A}, \mathcal{B}) \\ &= \frac{1}{|\mathcal{X}|}\Bigg[\sum_{x \in \mathcal{X}}{\Big[\sum_{a\in \mathcal{A}}{s(x,a)} - \sum_{b\in \mathcal{B}}{s(x,b)}\Big]} \\ &\hbox{\ \ \ \ \ \ \ \ \ \ \ \ } - \sum_{y \in \mathcal{Y}}{\Big[\sum_{a\in \mathcal{A}}{s(y,a)} - \sum_{b\in \mathcal{B}}{s(y,b)}\Big]}\Bigg],
\end{align*}
where $s(x,y) = \cos(\hbox{vec}(x), \hbox{vec}(y))$ and $\hbox{vec}(x) \in \mathbb{R}^k$ is the $k$-dimensional word embedding for word $x$. 
Note that for this definition of WEAT, the cardinality of the sets must be equal, so $|\mathcal{A}|=|\mathcal{B}|$ and $|\mathcal{X}|=|\mathcal{Y}|$. Our  conceptor formulation given below relaxes this assumption.

To motivate our conceptor formulation, we further generalize WEAT to capture the covariance between the target word and the attribute word embeddings. First, let  $X$, $Y$, $A$ and $B$ be matrices whose columns are word embeddings corresponding to the words in the sets $\mathcal{X}, \mathcal{Y}, \mathcal{A}, \mathcal{B}$, respectively (i.e. the two sets of target words and two sets of attribute words, respectively). To formally define this, without loss of generality choose $\mathcal{X}$, let $X = [x_i]_{i \in I}$ where for $i$ in an index set $I$ with cardinality $|\mathcal{X}|$ and $x_i = \hbox{vec}(x)$ where the word $x$ is indexed at the ith value of the index set.\footnote{To clarify, in our notation $x_i \in \mathbb{R}^k$ and $x \in \mathcal{X}$.} 
We can then write WEAT as,  
\[
\|X^TA - X^TB - (Y^TA - Y^TB)\|_F \]
\[
 = \|(X-Y)^T(A-B)\|_F,
\]
where $\|\cdot\|_F$ is the Frobenius norm. If the embeddings are unit length, then GWEAT is the same as $|\mathcal{X}|$ times WEAT.\footnote{Our generalization of WEAT is different from \citet{swinger2018what}.}

Suppose we want to mitigate bias by applying the $k \times k$ bias mitigating matrix, $G=\neg C$, which optimally removes bias from any matrix of word embeddings. We select $G$ to minimize
\[
  \|(G(X-Y))^TG(A-B)\|_F,
\]
\[
  = \|(X-Y)^TG^TG(A-B)\|_F.
\]
Since the conceptor, $C$, is calculated using the word embeddings of $\mathcal{Z}= \mathcal{X} \cup \mathcal{Y}$, the negated conceptor will mitigate the variance from the target sets, which hopefully identifies the most important bias directions.

\section{Embeddings}
For context-independent embeddings, we used off-the-shelf Fasttext subword embeddings\footnote{\url{https://dl.fbaipublicfiles.com/fasttext/vectors-english/crawl-300d-2M-subword.zip}.}, which were trained with subword information on the Common Crawl (600B tokens), the GloVe embeddings \footnote{\url{https://nlp.stanford.edu/projects/glove/}} trained on Wikipedia and Gigaword and word2vec\footnote{\url{}} trained on roughly 100 billion words from a Google News dataset. The embeddings used are not centered and normalized to unit length as in \citet{bolukbasi2016man}.

For contextualized embeddings, we used ELMo small which was trained on the 1 Billion Word Benchmark, approximately 800M tokens of news crawl data from WMT 2011.\footnote{\url{https://s3-us-west-2.amazonaws.com/allennlp/models/elmo/2x1024_128_2048cnn_1xhighway/elmo_2x1024_128_2048cnn_1xhighway_weights.hdf5}} We also experimented with the state-of-the-art contextual model ``BERT-Large, Uncased'' which has 24-layer, 1024-hidden, 16-heads, 340M parameters. BERT is trained on the BooksCorpus (0.8B words) and Wikipedia (2.5B words). We used the last four hidden layers of BERT.
We used the Brown Corpus for the word contexts to create instances of the ELMo and BERT embeddings. Embeddings of  English words only have been used for all the tests.

\begin{table*}[htb!]
    \centering
    \resizebox{0.95\textwidth}{!}{  
\begin{tabular}{|c|c|c|c|c|c|c|c|c|c|}
\hline
\multirow{2}{*}{Embedding} & \multicolumn{1}{c|}{\multirow{2}{*}{Subspace}} & \multicolumn{2}{c|}{Without Debiasing}        & \multicolumn{2}{c|}{Mu et al.} & \multicolumn{2}{c|}{Bolukbasi et al.} & \multicolumn{2}{c|}{Conceptor Negation} \\ \cline{3-10} 
                           & \multicolumn{1}{c|}{}                          & d                     & p                     & d              & p              & d                  & p                 & d                   & p                 \\ \hline
\multirow{5}{*}{Glove}     & Pronouns                                       & \multirow{5}{*}{1.78} & \multirow{5}{*}{0.00} & 1.81           & 0.00           & 1.24               & 0.01              & \textbf{0.13}                & 0.40              \\
                           & Extended List                                  &                       &                       & 1.86           & 0.00           & 1.24               & 0.01              & 0.36                & 0.26              \\
                           & Propernouns                                    &                       &                       & 1.74           & 0.00           & 1.24               & 0.01              & 0.78                & 0.07              \\
                           & All                                            &                       &                       & 1.75           & 0.00           & 1.20               & 0.01              & 0.35                & 0.27              \\
                           & OR                                            &                       &                       & NA             & NA             & NA                 & NA                & -0.51               & 0.81              \\ \hline
\multirow{5}{*}{word2vec}  & Pronouns                                       & \multirow{5}{*}{1.81} & \multirow{5}{*}{0.00} & 1.79           & 0.00           & 1.55               & 0.00              & 1.09                & 0.02              \\
                           & Extended List                                  &                       &                       & 1.79           & 0.00           & 1.59               & 0.00              & 1.38                & 0.00              \\
                           & Propernouns                                    &                       &                       & 1.70           & 0.0            & 1.59               & 0.0               & 1.45                & 0.00              \\
                           & All                                            &                       &                       & 1.71           & 0.00           & 1.56               & 0.00              & 1.40                & 0.00              \\
                           & OR                                            &                       &                       & NA             & NA             & NA                 & NA                & \textbf{0.84}                & 0.05              \\ \hline
\multirow{5}{*}{Fasttext}  & Pronouns                                       & \multirow{5}{*}{1.67} & \multirow{5}{*}{0.00} & 1.70           & 0.0            & 1.45               & 0.00              & 0.95                & 0.04              \\
                           & Extended List                                  &                       &                       & 1.70           & 0.0            & 1.47               & 0.00              & 0.84                & 0.04              \\
                           & Propernouns                                    &                       &                       & 0.86           & 0.06           & 1.47               & 0.00              & 0.85                & 0.06              \\
                           & All                                            &                       &                       & 0.82           & 0.05           & 1.14               & 0.01              & 0.81                & 0.06              \\
                           & OR                                            &                       &                       & NA             & NA             & NA                 & NA                & \textbf{0.24}                & 0.33              \\ \hline
\end{tabular}
}
\caption{Gender Debiasing non-contextualized embeddings: (Career, Family) vs (Male, Female)}
\label{table:career_family_we}
\end{table*}
\begin{table*}[htb!]
    \centering
\begin{tabular}{|c|c|c|c|c|c|c|c|}
\hline
\multirow{2}{*}{Embedding} & \multirow{2}{*}{Subspace} & \multicolumn{2}{l|}{Without Debiasing} & \multicolumn{2}{l|}{Mu et. al.} & \multicolumn{2}{l|}{Conceptor Negation} \\ \cline{3-8} 
 &  & \multicolumn{1}{c|}{d} & \multicolumn{1}{c|}{p} & \multicolumn{1}{c|}{d} & \multicolumn{1}{c|}{p} & \multicolumn{1}{c|}{d} & \multicolumn{1}{c|}{p} \\ \hline
\multicolumn{1}{|c|}{\multirow{5}{*}{ELMo}} & Pronouns & \multirow{5}{*}{1.79} & \multirow{5}{*}{0.0} & 1.79 & 0.00 & 0.70 & 0.10 \\ 
\multicolumn{1}{|c|}{} & Extended List &  &  & 1.79 & 0.00 & \textbf{0.06} & 0.46 \\
\multicolumn{1}{|c|}{} & Propernouns &  &  & 1.79 & 0.00 & -0.61 & 0.89 \\ 
\multicolumn{1}{|c|}{} & All &  &  & 1.79 & 0.00 & -0.28 & 0.73 \\ 
\multicolumn{1}{|c|}{} & OR &  &  & NA & NA & -0.85 & 0.96 \\ \hline
\multicolumn{1}{|c|}{\multirow{5}{*}{BERT}} & Pronouns & \multirow{5}{*}{1.21} & \multirow{5}{*}{0.01} & 1.21 & 0.01 & 1.31 & 0.00 \\ 
 & Extended List &  &  & 1.27 & 0.00 & 1.33 & 0.01 \\ 
 & Propernouns &  &  & 1.27 & 0.01 & 0.92 & 0.04 \\ 
 & All &  &  & 1.27 & 0.01 & \textbf{0.63} & 0.13 \\  
 & OR &  &  & NA & NA & 0.97 & 0.02 \\ \hline
\end{tabular}
\caption{Gender Debiasing Contextualized embeddings: (Career, Family) vs (Male, Female)}
\label{table:career_family_bert}
\end{table*}

\begin{table*}[htb!]
    \centering
    \resizebox{0.95\textwidth}{!}{  
\begin{tabular}{|c|c|c|c|c|c|c|c|c|c|}
\hline
\multirow{2}{*}{Embedding} & \multirow{2}{*}{Subspace} & \multicolumn{2}{l|}{Without Debiasing} & \multicolumn{2}{l|}{Mu et al.} & \multicolumn{2}{l|}{Bolukbasi et al} & \multicolumn{2}{l|}{Conceptor Negation} \\ \cline{3-10} 
 &  & \multicolumn{1}{c|}{d} & \multicolumn{1}{c|}{p} & \multicolumn{1}{c|}{d} & \multicolumn{1}{c|}{p} & \multicolumn{1}{c|}{d} & \multicolumn{1}{c|}{p} & \multicolumn{1}{c|}{d} & \multicolumn{1}{c|}{p} \\ \hline
\multirow{5}{*}{Glove} & Pronouns & \multirow{5}{*}{1.09} & \multirow{5}{*}{0.02} & 0.89 & 0.04 & -0.53 & 0.85 & 1.04 & 0.01 \\ 
 & Extended List &  &  & 1.07 & 0.02 & -0.60 & 0.86 & -0.52 & 0.83 \\ 
 & Propernouns &  &  & 1.04 & 0.02 & -0.56 & 0.86 & 0.20 & 0.33 \\ 
 & All &  &  & 1.03 & 0.02 & -0.53 & 0.82 & \textbf{0.18} & 0.35 \\ 
 & OR &  &  & NA & NA & NA & NA & -0.48 & 0.82 \\ \hline
\multirow{5}{*}{Word2vec} & Pronouns & \multirow{5}{*}{1.00} & \multirow{5}{*}{0.02} & 0.89 & 0.03 & -1.09 & 0.99 & 1.10 & 0.01 \\ 
 & Extended List &  &  & 1.00 & 0.03 & -1.14 & 1.00 & -0.49 & 0.82 \\ 
 & Propernouns &  &  & 0.88 & 0.04 & -1.17 & 1.00 & 0.33 & 0.27 \\ 
 & All &  &  & 0.90 & 0.04 & -1.07 & 0.99 & \textbf{0.25} & 0.34 \\ 
 & OR &  &  & NA & NA & NA & NA & -0.47 & 0.81 \\ \hline
\multicolumn{1}{|c|}{\multirow{5}{*}{Fasttext}} & Pronouns & \multirow{5}{*}{1.19} & \multirow{5}{*}{0.01} & 1.08 & 0.01 & 0.18 & 0.35 & -0.36 & 0.76 \\ 
\multicolumn{1}{|c|}{} & Extended List &  &  & 0.71 & 0.08 & 0.21 & 0.353 & 0.73 & 0.09 \\ 
\multicolumn{1}{|c|}{} & Propernouns &  &  & 0.12 & 0.43 & 0.15 & 0.40 & -0.47 & 0.80 \\ 
\multicolumn{1}{|c|}{} & All &  &  & \textbf{0.038} & 0.47 & 0.20 & 0.32 & -0.50 & 0.84 \\ 
\multicolumn{1}{|c|}{} & OR &  &  & NA & NA & NA & NA & -0.46 & 0.78 \\
\hline
\end{tabular}
}
\caption{Gender Debiasing non-contextualized embeddings: (Math, Arts) vs (Male, Female)}
\label{table:math_arts_we}
\end{table*}

\begin{table*}[htb!]
    \centering
\begin{tabular}{|c|c|c|c|c|c|c|c|}
\hline
\multirow{2}{*}{Embedding} & \multirow{2}{*}{Subspace} & \multicolumn{2}{l|}{Without Debiasing} & \multicolumn{2}{l|}{Mu et. al.} & \multicolumn{2}{l|}{Conceptor Negation} \\ \cline{3-8} 
 &  & \multicolumn{1}{c|}{d} & \multicolumn{1}{c|}{p} & \multicolumn{1}{c|}{d} & \multicolumn{1}{c|}{p} & \multicolumn{1}{c|}{d} & \multicolumn{1}{c|}{p} \\ \hline
\multicolumn{1}{|c|}{\multirow{5}{*}{ELMo}} & Pronouns & \multirow{5}{*}{0.94} & \multirow{5}{*}{0.02} & 0.94 & 0.03 & \textbf{-0.03} & 0.38 \\ 
\multicolumn{1}{|c|}{} & Extended List &  &  & 0.95 & 0.02 & 0.27 & 0.29 \\ 
\multicolumn{1}{|c|}{} & Propernouns &  &  & 0.94 & 0.02 & 0.85 & 0.05 \\ 
\multicolumn{1}{|c|}{} & All &  &  & 0.94 & 0.04 & 0.87 & 0.05 \\ 
\multicolumn{1}{|c|}{} & OR &  &  & NA & NA & 0.53 & 0.13 \\ \hline
\multicolumn{1}{|c|}{\multirow{5}{*}{BERT}} & Pronouns & \multirow{5}{*}{0.23} & \multirow{5}{*}{0.777} & 0.23 & 0.79 & 0.15 & 0.15 \\ 
 & Extended List &  &  & 0.16 & 0.82 & \textbf{0.06} & 0.53 \\ 
 & Propernouns &  &  & 0.16 & 0.82 & 0.75 & 0.08 \\
 & All &  &  & 0.16 & 0.85 & 0.43 & 0.24 \\ 
 & OR &  &  & NA & NA & -0.07 & 0.59 \\ \hline
\end{tabular}
\caption{Gender Debiasing contextualized embeddings: (Math, Arts) vs (Male, Female)}
\label{table:math_arts_bert}
\end{table*}

\begin{table*}[htb!]
    \centering
    \resizebox{0.95\textwidth}{!}{  
\begin{tabular}{|c|c|c|c|c|c|c|c|c|c|}
\hline
\multirow{2}{*}{Embedding} & \multirow{2}{*}{Subspace} & \multicolumn{2}{l|}{Without Debiasing} & \multicolumn{2}{l|}{Mu et al.} & \multicolumn{2}{l|}{Bolukbasi et al.} & \multicolumn{2}{l|}{Conceptor Negation} \\ \cline{3-10} 
 &  & d & p & d & p & d & p & d & p \\ \hline
\multirow{5}{*}{Glove} & Pronouns & \multirow{5}{*}{1.34} & \multirow{5}{*}{0.0} & 1.23 & 0.01 & -0.46 & 0.819 & \textbf{-0.20} & 0.66 \\ 
 & Extended List &  &  & 1.27 & 0.00 & -0.51 & 0.83 & 0.93 & 0.04 \\ 
 & Propernouns &  &  & 1.21 & 0.011 & -0.48 & 0.839 & 0.65 & 0.10 \\
 & All &  &  & 1.21 & 0.00 & -0.45 & 0.81 & 0.68 & 0.10 \\ 
 & OR &  &  & NA & NA & NA & NA & 0.60 & 0.12 \\ \hline
\multirow{5}{*}{Word2vec} & Pronouns & \multirow{5}{*}{1.16} & \multirow{5}{*}{0.01} & 1.09 & 0.02 & -0.46 & 0.80 & 0.45 & 0.21 \\ 
 & Extended List &  &  & 1.20 & 0.01 & -0.50 & 0.80 & 0.59 & 0.13 \\ 
 & Propernouns &  &  & 1.08 & 0.02 & -0.55 & 0.86 & 0.69 & 0.10 \\ 
 & All &  &  & 1.08 & 0.02 & -0.46 & 0.80 & 0.66 & 0.13 \\ 
 & OR &  &  & NA & NA & NA & NA & \textbf{0.09} & 0.45 \\ \hline
\multirow{5}{*}{Fasttext} & Pronouns & \multirow{5}{*}{1.48} & \multirow{5}{*}{0.00} & 1.51 & 0.00 & 0.88 & 0.04 & 0.93 & 0.03 \\ 
 & Extended List &  &  & 0.85 & 0.04 & 0.85 & 0.04 & 1.36 & 0.00 \\ 
 & Propernouns &  &  & 1.01 & 0.03 & 0.85 & 0.05 & \textbf{0.75} & 0.08 \\ 
 & All &  &  & 0.98 & 0.03 & 0.88 & 0.03 & 0.89 & 0.05 \\ 
 & OR &  &  & NA & NA & NA & NA & 0.89 & 0.05 \\
 \hline
\end{tabular}
}
\caption{Gender Debiasing non-cotextualized embeddings: (Science, Arts) vs (Male, Female)}
\label{table:science_arts_we}
\end{table*}
\begin{table*}[htb!]
    \centering
\begin{tabular}{|c|c|c|c|c|c|c|c|}
\hline
\multirow{2}{*}{Embedding} & \multirow{2}{*}{Subspace} & \multicolumn{2}{l|}{Without Debiasing} & \multicolumn{2}{l|}{Mu et. al.} & \multicolumn{2}{l|}{Conceptor Negation} \\ \cline{3-8} 
 &  & d & p & d & p & d & p \\ \hline
\multirow{5}{*}{ELMo} & Pronouns & \multirow{5}{*}{1.32} & \multirow{5}{*}{0.0} & 1.31 & 0.00 & \textbf{0.41} & 0.22 \\ 
 & Extended List &  &  & 1.32 & 0.005 & 0.52 & 0.24 \\ 
 & Propernouns &  &  & 1.38 & 0.00 & 1.28 & 0.00 \\ 
 & All &  &  & 1.34 & 0.00 & 0.92 & 0.03 \\ 
 & OR &  &  & NA & NA & 0.82 & 0.05 \\ \hline
\multirow{5}{*}{BERT} & Pronouns & \multirow{5}{*}{-0.91} & \multirow{5}{*}{0.88} & -0.91 & 0.87 & -1.23 & 0.97 \\ 
 & Extended List &  &  & -0.90 & 0.91 & -1.10 & 0.99 \\ 
 & Propernouns &  &  & -0.90 & 0.92 & -0.93 & 0.92 \\ 
 & All &  &  & -0.90 & 0.90 & \textbf{-0.38} & 0.70 \\ 
 & OR &  &  & NA & NA & 0.97 & 0.02 \\ \hline
\end{tabular}
\caption{Gender Debiasing cotextualized embeddings: (Science, Arts) vs (Male, Female)}
\label{table:science_arts_bert}
\end{table*}

\section{WEAT Debiasing Experiments}

As described in section~\ref{sec:motivationtheory}, WEAT assumes as its null hypothesis that there is no relative bias between the pair of concepts defined as the target words and attribute words. In our experiments, we measure the effect size (the WEAT score normalized by the standard deviation of differences of attribute words w.r.t target words) (d) and the one-sided p-value of the permutation test. A higher absolute value of effect size indicates larger bias between words in the target set with respect to the words in the attribute set. We would like the absolute value of the effect size to be zero. Since the p-value measures the likelihood that a random permutation of the attribute words would produce at least the observed test statistic, it should be high (at least 0.05) to indicate lack of bias in the positive direction.

Conceptually, the conceptor should be a soft projection matrix on the linear subspace representing the bias direction. For instance, the subspace representing gender must consist of words which are specific to or in some sense related to gender. 

A gender word list might be a set of pronouns which are specific to a particular gender such as \textit{he / she} or \textit{himself / herself} and gender specific words representing relationships like \textit{brother / sister} or \textit{uncle / aunt}. We test conceptor debiasing both using the list of such pronouns used by \citet{caliskan2017semantics} and using a more comprehensive list of gender-specific words that includes gender specific terms related to occupations, relationships and other commonly used words such as \textit{prince / princess} and \textit{host / hostess}\footnote{\url{https://github.com/uclanlp/corefBias}, \url{https://github.com/uclanlp/gn_glove}}. We further tested conceptor debiasing using male and female names such as \textit{Aaron / Alice} or \textit{Chris / Clary}\footnote{\url{https://www.cs.cmu.edu/Groups/AI/areas/nlp/corpora/names/}}. We also tested our method with the combination of all lists. The combination of the subspace was done in two ways - either by taking the union of all word lists or by applying the OR operator on the three conceptor matrices computed independently.

The subspace for racial bias was determined using list of European American and African American names.

We tested target pairs of Science vs. Arts,  Math vs. Arts, and
Career vs. Family word lists with the attribute of the male vs. female names to test gender debiasing. Similarly, we examined European American names vs. African American names as target pairs with the attribute of pleasant vs. unpleasant to test racial debiasing.

Our findings indicate that expanded lists give better debiasing for word embeddings; however, the results are not as clear for contextualized embeddings. The OR operator on two conceptors describing subspaces of pronouns/nouns and names generally outperforms a union of these words. This further motivates the use of the debiasing conceptor.

\subsection{Racial Debiasing Results}
\begin{table}[tbh]
\centering
\begin{tabular}{|l|c|c|c|c|}
\hline
\multirow{2}{*}{\textbf{Embedding}} & \multicolumn{2}{l|}{\textbf{Original}} & \multicolumn{2}{c|}{\textbf{\begin{tabular}[c]{@{}c@{}}Conceptor \\ Negation\end{tabular}}} \\ \cline{2-5} 
 & d & p & d & p \\ \hline
GloVe & 1.35 & 0.00 & 0.69 & 0.01 \\ \hline
word2vec & -0.27 & \textbf{0.27} & -0.55 & \textbf{0.72} \\ \hline
Fasttext & 0.41 & 0.04 & -0.27 & \textbf{0.57} \\ \hline
ELMo & 1.37 & 0.00 & -0.45 & \textbf{0.20} \\ \hline
BERT & 0.92 & 0.00 & 0.36 & \textbf{0.61} \\ \hline
\end{tabular}
\caption{Racial Debiasing: (European American Names, African American Names) vs (Pleasant, Unpleasant). d is the effect size, which we want to be close to $0$ and p is the p-value, which we want to be larger than 0.05.}
\label{tab:racialdebias}
\end{table}
Table \ref{tab:racialdebias} summarizes the effect size (d) and the one-sided p-value we obtained by running WEAT on each of the word embeddings for racial debiasing. In this experiment we used the same setup as \citet{caliskan2017semantics} and compare attribute Words of European American / African American names with target words ``pleasant'' and ``unpleasant''.
In Table \ref{tab:racialdebias} we see that racial bias is mitigated in all cases aside from GloVe. Furthermore, for word2vec the associational bias is not significant. 
We also found that the conceptor nearly always outperforms the hard debiasing methods of \citet{Mu2018} and \citet{bolukbasi2016man}. 

\subsection{Gender Debiasing Results}

Tables \ref{table:career_family_we}, \ref{table:math_arts_we} and \ref{table:science_arts_we} show the results obtained on gender debiasing between attribute words of ``Family'' and ``Career', ``Math'' and ``Arts'' and ``Science'' and ``Arts'' with the target words ``Male'' and ``Female'' respectively for the traditional word embeddings. We show the results for all the word representations; however, the method of \citet{bolukbasi2016man} can only be applied to standard word embeddings.\footnote{The concurrent work of \citet{wang2019genderbiaselmo} was not available in time for us to compare with this method.} We show the results when embeddings are debiased using conceptors computed using different subspaces. It can be seen in the tables that the bias for the conceptor negated embeddings is significantly less than that of the original embeddings. In the tables, the conceptor debiasing method is compared with the hard-debiasing technique proposed by \citet{Mu2018} where the first principal component of the subspace from the embeddings is completely project off. The debiasing conceptor outperforms the hard debiasing technique in almost all cases. Note that the OR operator can not be used with the hard debiasing technique and thus is not reported.

Similarly, Tables \ref{table:career_family_bert}, \ref{table:math_arts_bert} and \ref{table:science_arts_bert} show a comparison of the effect size and  p-value using the hard debiasing technique and conceptor debiasing on conceptualized embeddings. It can be seen that conceptor debiasing generally outperforms other methods in mitigating (has a small absolute value) bias with the ELMo embeddings for all the subspaces. The results are less clear for BERT as observed in Table \ref{table:science_arts_bert}, which we will discuss in the following section. Note that combining all subspaces gives a significant reduction in the effect size.

\subsection{Discussion of BERT Results}

One of our most surprising findings is that unlike ELMo, the bias in BERT according to WEAT is less consistent than other word representations;  WEAT effect sizes in BERT vary largely across different layers. Furthermore, the debiasing conceptor occasionally creates reverse bias in BERT, suggesting that tuning of the hyper-parameter $\alpha$ may be required. Another possibility is that BERT is capturing multiple concepts, and the presumption that the target lists are adequately capturing gender or racial attributes is incorrect. This suggests that further study into word lists is called for, along with  visualization and end-task evaluation. 
It should also be noted that our results are in line with those from \citet{may2019measuring}. 

\section{Retaining Semantic Similarity}
In order to understand if the debiasing conceptor was harming the semantic content of the word embeddings, we examined conceptor debiased embedding for semantic similarity tasks. As done in \citet{Liu2018correcting} we used the seven standard word similarity test set and report Pearson's correlation. The word similarity sets are: the RG65 \citep{Rubenstein1965},  the WordSim-353 (WS) \citep{Finkelstein2002},  the rare-words (RW) \citep{Luong2013}, the MEN dataset \citep{Bruni2014}, the MTurk \citep{Radinsky2011}, the SimLex-999 (SimLex) \citep{Hill2015}, and the SimVerb-3500 \citep{Gerz2016}. Table~\ref{tab:semsimtasks} shows that conceptors help in preserving and at times increasing the semantic information in the embeddings. It should be noted that these tasks can not be applied to contextualized embeddings such as ELMo and BERT. So, we do not report these results.

\begin{table}[h!]
\resizebox{\columnwidth}{!}{%
\begin{tabular}{lllllll}
 & \multicolumn{2}{c}{GloVe} & \multicolumn{2}{c}{word2vec} & \multicolumn{2}{c}{Fasttext} \\ \cline{2-7} 
 & Orig. & CN & Orig. & CN & Orig. & CN \\ \hline
RG65 & \textbf{76.03}  & 70.92  & 74.94  & \textbf{78.58}  & 85.87  & \textbf{85.94} \\
WS & 73.79  & \textbf{75.17}  & 69.34  & \textbf{69.34}  & \textbf{78.82}  & 77.44  \\
RW & 51.01  & \textbf{55.25}  & 55.78  & \textbf{56.04}  & 62.17 & \textbf{62.48 } \\
MEN & \textbf{80.13}  & 80.10  & 77.07  & \textbf{77.85}  & \textbf{83.64} & 82.64  \\
MTurk & 69.16  & \textbf{71.17} & \textbf{68.31}  & 67.68  & \textbf{72.45} & 71.34  \\
SimLex & 40.76  & \textbf{45.85}  & 44.27  & \textbf{46.05}  & 50.55  & \textbf{50.78}  \\
SimVerb & 28.42 & \textbf{34.51} & 36.54 & \textbf{37.33}  & \textbf{42.75} & 42.72 \\ \hline
\end{tabular}
}
\caption{Word Similarity comparison with conceptor debiased embeddings using all gender words as conceptor subspace.}
\label{tab:semsimtasks}
\end{table}

\section{Conclusion}
We have shown that the debiasing conceptor can  successfully debias word embeddings, outperforming previous state-of-the art 'hard' debiasing methods.
Best results are obtained when lists are broken up into subsets of 'similar' words (pronouns, professions, names, etc.), and separate conceptors are learned for each subset and then OR'd.  Conceptors for different protected subclasses such as gender and race can be similarly OR'd to jointly debias.

Contextual embeddings such as ELMo and BERT, which give a different vector for each word token, work particularly well with conceptors, since they produce a large number of embeddings; however, further research on tuning conceptors for BERT needs to be done. Finally, we note that embedding debiasing may leave bias which is undetected by measures such as WEAT \citet{gonen2019lipstick}; thus, all debiasing methods should be tested on end-tasks such as emotion classification and co-reference resolution.

\bibliography{ref}

\begin{thebibliography}{32}
\expandafter\ifx\csname natexlab\endcsname\relax\def\natexlab#1{#1}\fi

\bibitem[{Bhatia(2017)}]{bhatia2017semantic}
Sudeep Bhatia. 2017.
\newblock The semantic representation of prejudice and stereotypes.
\newblock \emph{Cognition}, 164:46--60.

\bibitem[{Bolukbasi et~al.(2016)Bolukbasi, Chang, Zou, Saligrama, and
  Kalai}]{bolukbasi2016man}
Tolga Bolukbasi, Kai-Wei Chang, James~Y Zou, Venkatesh Saligrama, and Adam~T
  Kalai. 2016.
\newblock Man is to computer programmer as woman is to homemaker? debiasing
  word embeddings.
\newblock In \emph{Advances in neural information processing systems}, pages
  4349--4357.

\bibitem[{Bruni et~al.(2014)Bruni, Tran, and Baroni}]{Bruni2014}
E.~Bruni, N.~K. Tran, and M.~Baroni. 2014.
\newblock Multimodal distributional semantics.
\newblock \emph{Journal of Artificial Intelligence Research}, 49(1):1--47.

\bibitem[{Caliskan et~al.(2017)Caliskan, Bryson, and
  Narayanan}]{caliskan2017semantics}
Aylin Caliskan, Joanna~J Bryson, and Arvind Narayanan. 2017.
\newblock Semantics derived automatically from language corpora contain
  human-like biases.
\newblock \emph{Science}, 356(6334):183--186.

\bibitem[{Dixon et~al.(2018)Dixon, Li, Sorensen, Thain, and
  Vasserman}]{dixon2018measuring}
Lucas Dixon, John Li, Jeffrey Sorensen, Nithum Thain, and Lucy Vasserman. 2018.
\newblock Measuring and mitigating unintended bias in text classification.
\newblock In \emph{Proceedings of the 2018 AAAI/ACM Conference on AI, Ethics,
  and Society}, pages 67--73. ACM.

\bibitem[{Finkelstein et~al.(2002)Finkelstein, Gabrilovich, Matias, Rivlin,
  Solan, Wolfman, and Ruppin}]{Finkelstein2002}
L.~Finkelstein, E.~Gabrilovich, Y.~Matias, E.~Rivlin, Z.~Solan, G.~Wolfman, and
  E.~Ruppin. 2002.
\newblock Placing search in context: the concept revisited.
\newblock \emph{ACM Transactions on Information Systems}, 20(1):116--131.

\bibitem[{Friedler et~al.(2016)Friedler, Scheidegger, and
  Venkatasubramanian}]{friedler2016possibility}
Sorelle~A Friedler, Carlos Scheidegger, and Suresh Venkatasubramanian. 2016.
\newblock On the (im) possibility of fairness.
\newblock \emph{arXiv preprint arXiv:1609.07236}.

\bibitem[{Gerz et~al.(2016)Gerz, Vulic, Hill, Reichart, and
  Korhonen}]{Gerz2016}
D.~Gerz, I.~Vulic, F.~Hill, R.~Reichart, and A.~Korhonen. 2016.
\newblock {S}im{V}erb-3500: a large-scale evaluation set of verb similarity.
\newblock In \emph{Proceedings of the {EMNLP} 2016}, pages 2173--2182.

\bibitem[{Gonen and Goldberg(2019)}]{gonen2019lipstick}
Hila Gonen and Yoav Goldberg. 2019.
\newblock Lipstick on a pig: Debiasing methods cover up systematic gender
  biases in word embeddings but do not remove them.
\newblock In \emph{North American Chapter of the Association for Computational
  Linguistics (NAACL)}.

\bibitem[{Greenwald and Banaji(1995)}]{greenwald1995implicit}
Anthony~G Greenwald and Mahzarin~R Banaji. 1995.
\newblock Implicit social cognition: attitudes, self-esteem, and stereotypes.
\newblock \emph{Psychological review}, 102(1):4.

\bibitem[{Greenwald et~al.(1998)Greenwald, McGhee, and
  Schwartz}]{greenwald1998measuring}
Anthony~G Greenwald, Debbie~E McGhee, and Jordan~LK Schwartz. 1998.
\newblock Measuring individual differences in implicit cognition: the implicit
  association test.
\newblock \emph{Journal of personality and social psychology}, 74(6):1464.

\bibitem[{He and Jaeger(2018)}]{He2018}
X.~He and H.~Jaeger. 2018.
\newblock \href {https://openreview.net/pdf?id=B1al7jg0b} {Overcoming
  catastrophic interference using conceptor-aided backpropagation}.
\newblock In \emph{International Conference on Learning Representations}.

\bibitem[{Hill et~al.(2015)Hill, Reichart, and Korhonen}]{Hill2015}
F.~Hill, R.~Reichart, and A.~Korhonen. 2015.
\newblock Simlex-999: Evaluating semantic models with (genuine) similarity
  estimation.
\newblock \emph{Computational Linguistics}, 41(4):665--695.

\bibitem[{Hovy(2015)}]{hovy2015demographic}
Dirk Hovy. 2015.
\newblock Demographic factors improve classification performance.
\newblock In \emph{Proceedings of the 53rd Annual Meeting of the Association
  for Computational Linguistics and the 7th International Joint Conference on
  Natural Language Processing (Volume 1: Long Papers)}, volume~1, pages
  752--762.

\bibitem[{Hovy and Spruit(2016)}]{hovy2016social}
Dirk Hovy and Shannon~L Spruit. 2016.
\newblock The social impact of natural language processing.
\newblock In \emph{Proceedings of the 54th Annual Meeting of the Association
  for Computational Linguistics (Volume 2: Short Papers)}, volume~2, pages
  591--598.

\bibitem[{Jaeger(2014)}]{Jaeger2014}
H.~Jaeger. 2014.
\newblock \href {https://arxiv.org/pdf/1403.3369.pdf} {Controlling recurrent
  neural networks by conceptors}.
\newblock Technical report, Jacobs University Bremen.

\bibitem[{Kiritchenko and Mohammad(2018)}]{kiritchenko2018examining}
Svetlana Kiritchenko and Saif Mohammad. 2018.
\newblock Examining gender and race bias in two hundred sentiment analysis
  systems.
\newblock In \emph{Proceedings of the Seventh Joint Conference on Lexical and
  Computational Semantics}, pages 43--53.

\bibitem[{Liu et~al.(2018)Liu, Sedoc, and Ungar}]{Liu2018correcting}
T.~Liu, J.~Sedoc, and L.~Ungar. 2018.
\newblock Correcting the common discourse bias in linear representation of
  sentences using conceptors.
\newblock In \emph{Proceedings of ACM-BCB- 2018 Workshop on BioCreative/OHNLP
  Challenge, Washington, D.C., 2018}.

\bibitem[{Liu et~al.(2019{\natexlab{a}})Liu, Ungar, and
  Sedoc}]{liu2019continual}
T.~Liu, L.~Ungar, and J.~Sedoc. 2019{\natexlab{a}}.
\newblock Continual learning for sentence representations using conceptors.
\newblock In \emph{Proceedings of the {NAACL} {HLT} 2019}.

\bibitem[{Liu et~al.(2019{\natexlab{b}})Liu, Ungar, and Sedoc}]{Liu2019}
T.~Liu, L.~Ungar, and J.~Sedoc. 2019{\natexlab{b}}.
\newblock \href {https://arxiv.org/pdf/1811.11001.pdf} {Unsupervised
  post-processing of word vectors via conceptor negation}.
\newblock In \emph{Proceedings of the Thirty-Third {AAAI} Conference on
  Artificial Intelligence (AAAI-2019), Honolulu}.

\bibitem[{Lu et~al.(2018)Lu, Mardziel, Wu, Amancharla, and
  Datta}]{lu2018gender}
Kaiji Lu, Piotr Mardziel, Fangjing Wu, Preetam Amancharla, and Anupam Datta.
  2018.
\newblock Gender bias in neural natural language processing.
\newblock \emph{arXiv preprint arXiv:1807.11714}.

\bibitem[{Luong et~al.(2013)Luong, Socher, and Manning}]{Luong2013}
M.~Luong, R.~Socher, and C.~D. Manning. 2013.
\newblock Better word representations with recursive neural networks for
  morphology.
\newblock In \emph{Proceedings of the {CoNLL} 2013}.

\bibitem[{May et~al.(2019)May, Wang, Bordia, Bowman, and
  Rudinger}]{may2019measuring}
Chandler May, Alex Wang, Shikha Bordia, Samuel~R Bowman, and Rachel Rudinger.
  2019.
\newblock On measuring social biases in sentence encoders.
\newblock In \emph{North American Chapter of the Association for Computational
  Linguistics (NAACL)}.

\bibitem[{Mu and Viswanath(2018)}]{Mu2018}
J.~Mu and P.~Viswanath. 2018.
\newblock All-but-the-top: Simple and effective postprocessing for word
  representations.
\newblock In \emph{International Conference on Learning Representations}.

\bibitem[{Nikhil~Garg and Zou(2018)}]{garg2018}
Dan~Jurafsky Nikhil~Garg, Londa~Schiebinger and James Zou. 2018.
\newblock Word embeddings quantify 100 years of gender and ethnic stereotypes.
\newblock \emph{PNAS}.

\bibitem[{Radinsky et~al.(2011)Radinsky, Agichtein, Gabrilovich, and
  Markovitch}]{Radinsky2011}
K~Radinsky, E.~Agichtein, E.~Gabrilovich, and S.~Markovitch. 2011.
\newblock A word at a time: Computing word relatedness using temporal semantic
  analysis.
\newblock In \emph{Proceedings of the 20th International World Wide Web
  Conference}, pages 337--346, Hyderabad, India.

\bibitem[{Rubenstein and Goodenough(1965)}]{Rubenstein1965}
H.~Rubenstein and J.~B. Goodenough. 1965.
\newblock Contextual correlates of synonymy.
\newblock \emph{Communications of the ACM}, 8(10):627--633.

\bibitem[{Rudinger et~al.(2018)Rudinger, Naradowsky, Leonard, and
  Van~Durme}]{rudinger2018gender}
Rachel Rudinger, Jason Naradowsky, Brian Leonard, and Benjamin Van~Durme. 2018.
\newblock Gender bias in coreference resolution.
\newblock \emph{arXiv preprint arXiv:1804.09301}.

\bibitem[{Swinger et~al.(2018)Swinger, De{-}Arteaga, IV, Leiserson, and
  Kalai}]{swinger2018what}
Nathaniel Swinger, Maria De{-}Arteaga, Neil Thomas~Heffernan IV, Mark D.~M.
  Leiserson, and Adam~Tauman Kalai. 2018.
\newblock \href {http://arxiv.org/abs/1812.08769} {What are the biases in my
  word embedding?}
\newblock \emph{CoRR}, abs/1812.08769.

\bibitem[{Wang et~al.(2019)Wang, Zhao, Yatskar, Cotterell, Ordonez, and
  Chang}]{wang2019genderbiaselmo}
Tianlu Wang, Jieyu Zhao, Mark Yatskar, Ryan Cotterell, Vicente Ordonez, and
  Kai-Wei Chang. 2019.
\newblock Gender bias in contextualized word embeddings.
\newblock In \emph{North American Chapter of the Association for Computational
  Linguistics (NAACL)}.

\bibitem[{Yatskar et~al.(2016)Yatskar, Zettlemoyer, and
  Farhadi}]{yatskar2016situation}
Mark Yatskar, Luke Zettlemoyer, and Ali Farhadi. 2016.
\newblock Situation recognition: Visual semantic role labeling for image
  understanding.
\newblock In \emph{Proceedings of the IEEE Conference on Computer Vision and
  Pattern Recognition}, pages 5534--5542.

\bibitem[{Zhao et~al.(2017)Zhao, Wang, Yatskar, Ordonez, and
  Chang}]{zhao2017men}
Jieyu Zhao, Tianlu Wang, Mark Yatskar, Vicente Ordonez, and Kai-Wei Chang.
  2017.
\newblock Men also like shopping: Reducing gender bias amplification using
  corpus-level constraints.
\newblock In \emph{Proceedings of the 2017 Conference on Empirical Methods in
  Natural Language Processing}, pages 2979--2989.

\end{thebibliography}
\bibliographystyle{acl_natbib}

\appendix

\end{document}